\pdfoutput=1
\documentclass[11pt,a4paper]{article}
\usepackage[hyperref]{emnlp2020}
\usepackage{times}
\usepackage{latexsym}

\usepackage[utf8]{inputenc}

\usepackage{microtype}

\aclfinalcopy 

\newif\ifwithappendix\withappendixtrue

\newif\ifappendixshown
\appendixshowntrue

\usepackage[T1]{fontenc}
\usepackage[utf8]{inputenc}
\usepackage[textsize=tiny]{todonotes}
\usepackage{graphicx}
\usepackage{booktabs}
\usepackage{latexsym}

\usepackage{footnote}
\makesavenoteenv{tabular}
\makesavenoteenv{table}
\makesavenoteenv{table*}

\usepackage{xurl}

\newcommand\minput[1]{%
  \input{#1}%
  \ifhmode\ifnum\lastnodetype=11 \unskip\fi\fi}

\usepackage{hyperref}
\usepackage{xstring}

\newcommand{\code}[1]{\texttt{#1}}
\newcommand{\noqa}[1]{}
\newcommand{\noqall}[1]{}

\usepackage{amsmath}
\usepackage{longtable}
\usepackage{makecell}
\usepackage{bm}
\usepackage{booktabs}
\usepackage{float}
\usepackage{enumitem}
\usepackage{comment}

\title{Contract Discovery: Dataset and a Few-Shot Semantic Retrieval Challenge with Competitive Baselines}

\author{
 Łukasz Borchmann \and Dawid Wiśniewski \and Andrzej Gretkowski \\
 {\bf Izabela Kosmala} \and {\bf Dawid Jurkiewicz} \and {\bf Łukasz Szałkiewicz} \\
 {\bf Gabriela Pałka} \and {\bf Karol Kaczmarek} \and {\bf Agnieszka
 Kaliska} \and {\bf Filip Graliński} \\
 \\
 Applica.ai, Warsaw, Poland \\
  {\tt \{firstname.surname\}@applica.ai} \\
}

\begin{document}
\maketitle
\begin{abstract}
We propose a~new shared task of semantic retrieval from legal texts, in which a~so-called \textit{contract discovery} is to be performed--where legal clauses are extracted from documents, given a~few examples of similar clauses from other legal acts. The task differs substantially from conventional NLI and shared tasks on legal information extraction (e.g., one has to identify text span instead of a single document, page, or paragraph). The specification of the proposed task is followed by an evaluation of multiple solutions within the unified framework proposed for this branch of methods. It is shown that state-of-the-art pretrained encoders fail to provide satisfactory results on the task proposed. In contrast, Language Model-based solutions perform better, especially when unsupervised fine-tuning is applied. Besides the ablation studies, we addressed questions regarding detection accuracy for relevant text fragments depending on the number of examples available. In addition to the dataset and reference results, LMs specialized in the legal domain were made publicly available.

\end{abstract}

\hyphenation{Pe\~{n}a-ga-rikano}\noqa{spell-ga}\noqa{spell-rikano}

\noqall{spell-SNLI,spell-OUSA,spell-DCT,spell-ActivityNet,spell-MultiNLI}

\section{Introduction}

Processing of legal contracts requires significant human resources due to the complexity of documents, the expertise required and the consequences\noqa{grammar-CONSEQUENCES_OF_FOR} at stake. 
Therefore, a~lot of effort has been made to automate such tasks in order to limit processing costs--notice that law was one of the first areas where electronic information retrieval systems were adopted~\cite{Maxwell2008ConceptAC}.

Enterprise solutions referred to as \textit{contract discovery} deal with tasks, such as ensuring the inclusion of relevant clauses or their retrieval for further analysis (e.g.,~risk assessment). Such processes can consist of a~manual definition of a~few examples, followed by conventional information retrieval. This approach was taken recently by~\citet{DBLP:journals/corr/abs-1809-04262} for the extraction of fairness policies spread across agreements and administrative regulations.


\begin{figure*}
    \centering
    \includegraphics[width=\textwidth]{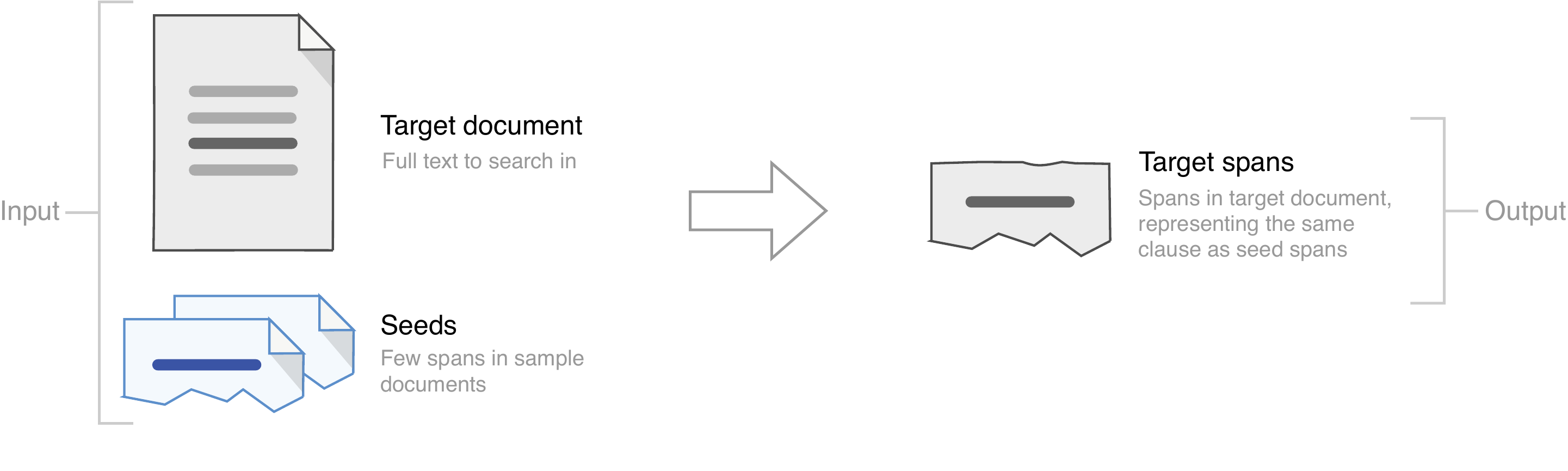}
    \caption{The aim of this task is to identify spans in the requested documents (referred to as \textit{target} documents) representing clauses analogous to the spans selected in other documents (referred to as \textit{seed} documents).\label{fig:idea}}
\end{figure*}

\begin{table}
    \small\centering
    \begin{tabular}{llccc}
        \toprule
        \textbf{Task} &
        \textbf{Legal} &
        \textbf{SI} &
        \textbf{Few-shot} \\
        \midrule
        COLIEE & $+$ & $-$ & $-$ \\
        SNLI & $-$ & $-$ & $-$ \\
        MultiNLI & $-$ & $-$ & $-$ \\
        TREC\noqa{spell-TREC} Legal Track & $+$ & $-$ & $-$ \\
        Propaganda detection & $-$ & $+$ & $-$ \\
        THUMOS\noqa{spell-THUMOS} (video) & $-$ & $+$ & $+$ \\
        ActivityNet (video) & $-$ & $+$ & $+$ \\
        ALBAYZIN\noqa{spell-ALBAYZIN} (audio) & $-$ & $+$ & $-$ \\
        \midrule
        Contract Discovery (ours) & $+$ & $+$ & $+$ \\
        \bottomrule
    \end{tabular}
    \caption{Comparison of existing shared tasks. Most of the related NLP tasks do not assume Span Identification (\textit{SI}), even those outside the legal domain (\textit{Legal}). Moreover, the few-shot setting is not popular within the field of NLP yet.\label{tab:comparison}}
\end{table}

\section{Review of Existing Datasets}

Table~\ref{tab:comparison} summarizes main differences between available challenges. It is shown that most of the related NLP tasks do not assume span identification, even those outside the legal domain. Moreover, the few-shot setting is not popular within the field of NLP yet.

None of existing tasks involving semantic similarity methods, such as SNLI~\cite{snli:emnlp2015} or multi-genre NLI~\cite{snli:emnlp2015}, assume span identification. Instead, standalone sentences are provided to determine their entailment. It is also the case of existing shared tasks for legal information extraction, such as COLIEE~\cite{Kano2017OverviewOC}, where one has to recognize entailment between articles and queries, as considered in the question answering problem. Obviously, the tasks aimed at retrieving documents consisting of multiple sentences, such as TREC\noqa{spell-TREC} legal track~\cite{Baron06trec-2006legal, oard2010evaluation, Chu2011FactorsAR}, lack this component.

There are a few NLP tasks where span identification is performed. These include some of plagiarism detection competitions~\cite{potthast-etal-2010-evaluation} and recently introduced SemEval\noqa{spell-SemEval} task of propaganda techniques detection~\cite{DaSanMartinoSemeval20task11}. When different media are considered, NLP span identification task is equivalent to the action recognition in temporally untrimmed videos where one is expected to provide the start and end times for detected activity. These include \noqa{spell-THUMOS}THUMOS 14~\cite{THUMOS14} as well as ActivityNet 1.2 and ActivityNet 1.3 challenges~\cite{caba2015activitynet}. Another example is query-by-example spoken term detection, as considered \noqa{spell-ALBAYZIN}e.g., ~in ALBAYZIN 2018 challenge~\cite{10.1186/s13636-019-0156-x}.

In a~typical business case of \textit{contract discovery} one may expect only a minimal number of examples. The number of available annotations results from the fact that \textit{contract discovery} is performed constantly for different clauses, and it is practically impossible to prepare data in a~number required by a conventional classifier every time. When one is interested in the few-shot setting, especially querying by multiple examples, there are no similar \noqa{grammar-ADVERB_OR_HYPHENATED_ADJECTIVE}shared tasks within the field of NLP. Some authors however experimented recently with few-shot Named Entity Recognition~\cite{Fritzler:2019:FCN:3297280.3297378} or few-shot text classification~\cite{bao2019few}. The first, however, involves identification of short spans (from one to few words), whereas the second does not assume span identification at all.

What is important, existing tasks aimed at recognizing textual entailment in natural language~\cite{snli:emnlp2015}, differ in terms of the domain. This also applies to a~multi-genre NLI~\cite{Williams2017ABC}, since legal texts vary significantly from other genres. As it will be shown later, methods optimal for MultiNLI do not perform well on the proposed task.

\section{Contract Discovery: New Dataset and Shared Task}\label{sec:materials}


In this section, we introduce a new dataset of \textit{Contract Discovery}, as
well as a derived few-shot semantic retrieval shared task.

\subsection{Desiderata}

We define our desiderata as follows. We wish to construct a dataset for testing the mechanisms that detect various types of regulations in legal documents. Such systems should be able to process unstructured text; that is, no legal documents segmentation into the hierarchy of distinct (sub)sections is to be given in advance. In other words, we want to provide natural language streams lacking formal structure, as in most of the real-word usage scenarios \cite{Vanderbeck2011AML}. What is more, it is assumed that a searched passage can be any part of the document and not necessarily a complete paragraph, subparagraph, or a clause. Instead, the process should be considered as a span identification task.

We intend to develop a dataset for identifying spans in a query-by-example scenario instead of the setting where articles are being returned as an answer for the question specified in natural language.

We wish to propose using this dataset in a few-shot scenarios, where one queries the system using multiple examples rather than a single one. The intended form of the challenge following these requirements is presented in Figure~\ref{fig:idea}. Roughly speaking, the task is to identify spans in the requested documents (referred to as \textit{target} documents) representing clauses analogous (i.e.~semantically and functionally equivalent) to the examples provided in other documents (referred to as \textit{seed} documents).


\subsection{Data Collection and Annotation}

Random subsets of bond issue prospectuses and non-disclosure agreement documents from the US EDGAR database\footnote{\url{http://www.www.sec.gov/edgar.shtml}}, as well as annual reports of charitable organizations from the UK Charity Register\footnote{\url{http://www.gov.uk/find-charity-information}} were annotated. Note there are no copyright issues and both datasets belong to the public domain.

Annotation was performed in such a~way that clauses of the same type were selected (e.g.,~determining the governing law, merger restrictions, tax changes call, or reserves policy). Clause types depend on the type of a legal act and can consist of a~single sentence, multiple sentences or sentence fragments. The exact type of a clause is not important\noqa{proselint-strunk_white.composition} during the evaluation since no full-featured training is allowed and a~set of only a~few sample clauses can be used during execution.

We restricted ourselves to 21 types as a result of a trade-off between annotation cost and the ability to formulate general remarks. Note that each clause type must be well-understood by the annotator (we described each very carefully in the instructions), and one must have all of the considered clauses in mind when the legal acts are being read during the process. In real-world legal applications, the clauses change in an everyday manner and depend on the problem analyzed by the layer at the moment.

Each document was annotated by two experts, and then reviewed (or resolved) by a~super-annotator, who also decided the gold standard. An average Soft $F_1$ score (Section~\ref{sec:evaluation}) of the two primary annotators, when compared to the gold standard (after the super-annotation), was taken to estimate human baseline performance of $0.84$.

The inter-annotator agreement was equal to $0.76$ in terms of Soft $F_1$ metric (Section~\ref{sec:evaluation}). It should be treated as an agreement between two randomly picked annotations since the total number of annotators was 10 (annotators were aligned randomly to a subset of documents in such a way that there would be two annotations and super-annotation per document).

Table~\ref{tab:clauses3} presents examples of clauses annotated in the sub-group of Charity Annual Reports documents. The detailed list of clauses and their examples can be found in Appendix~C.

The dataset is made publicly available. In addition, we release a~large, cleaned, plain-text corpus of legal and financial texts for the purposes of unsupervised model training or fine-tuning. 
All the available documents of US EDGAR as for November 19, 2018 were crawled. The resulting corpus consists of approx.~1M documents and 2B words in total (1.5G of text after \code{xz} compression).

\subsection{Core Statistics}

More than 2,500 spans were annotated in around 600 documents representing either bond issue prospectuses, non-disclosure agreement documents or annual reports of charitable organizations (the detailed statistics regarding the dataset are presented in Table~\ref{tab:stats}).

Annotated clauses differ substantially from what can be found in existing sentence entailment challenges in terms of sentence length and complexity. SNLI contains less than 1\% of sentences longer than 20 words, MultiNLI 5\%, whereas in the case of clauses, we expect to return and consider it is 93\% (and 77\% of all spans in our shared task are longer than 20 words).


\subsection{Evaluation Framework}

Documents were split into halves to form validation and test sets for the purposes of few-shot semantic retrieval challenge. Evaluation is performed by means of a~repeated random sub-sampling validation procedure. Sub-samples ($k$-combinations for each of 21 clauses, $k \in [2, 6]$) drawn from a~particular set of annotations are split into $k-1$ \textit{seed} documents and $1$ \textit{target} document. Thus, clauses similar to the \textit{seed} are expected to be returned from the target. We observed that the choice of input examples have an immense impact on the score. It is thus far more important to evaluate various \textit{seed} configurations that various target documents. On the other hand, we wanted to keep the computational cost of evaluation reasonably small, so either the number of seed configurations had to be reduced or the number of target documents for each configuration.

The selected $k$ interval results in 1-shot to 5-shot learning, considered to be few-shot learning~\cite{Wang2019GeneralizingFA}, whereas with the chosen number of sub-samples we expect improvements of 0.01 $F_1$ to be significant. Note that the 1–5 range denotes the number of annotated documents available, and it is possible that the same clause type appeared twice in one document, resulting in a~higher number of clause instances.

Soft $F_{1}$ metric on character-level spans is used for the purpose of evaluation, as implemented in \textit{GEval} tool~\cite{gralinski-etal-2019-geval}. Roughly speaking, this is the conventional $F_{1}$ measure, with precision and recall definitions altered to reflect the partial success of returning entities. In the case of the expected clause ranging between $[1, 4]$ characters and the answer with ranges $[1, 3]$, $[10, 15]$ (the system assumes a~clause occurs twice within the document), recall equals $0.75$ (since this is the part of the relevant item selected) and precision equals ca.~$0.33$ (since this is the number of selected characters which turned out to be relevant). The Hungarian algorithm \cite{c225e96c4acd499eb7a5ceabd178a543} is employed to solve the problem of expected and returned range assignments. Soft $F_{1}$ has the desired property of being based on the widely utilized $F_{1}$ metric while abandoning the binary nature of the match, which is undesirable in the case dealt with in the task described.

\begin{table}[tbp] 
\small
\centering
\begin{tabular}{lr}
\toprule
\textbf{Statistic} \\
\midrule
Documents annotated & 586 \\ 
Mean document length (words) & 24,284 \\
Clause types & 21 \\
Mean clause length (words) & 110 \\
Clause instances & 2,663 \\
\bottomrule
\end{tabular}
\caption{Core statistics regarding released dataset.}\label{tab:stats}
\end{table}
{
    \begingroup
    \begin{table*}[t]
    \small
    \begin{tabular}{p{0.37\textwidth} p{0.57\textwidth}}
    \toprule
    \textbf{Clause (Instances)} & \textbf{Example} \\
    \midrule
    \textsc{Main Objective (195/231)} The main objective of a charitable organization. & The aim of the Scout Association is to promote the development of young people in achieving their full physical, intellectual, social and spiritual potentials, as individuals, as responsible citizens and as members of their local, national and international communities. The method of achieving the Aim of the Association is by providing an enjoyable and attractive scheme of progressive training based on the Scout Promise and Law and guided by Adult leadership. \\\midrule
    \textsc{Governing Document (160/174)} Information about the legal document which represents the rule book for the way in which a charity operates (title, date of creation etc.). & The Open University Students Educational Trust (OUSET)\noqa{spell-OUSET} is controlled by its governing document, a~deed of trust, dated 22 May 1982 as amended by a~scheme dated 9 October 1992 and constitutes an unincorporated charity. \\\midrule
    \textsc{Trustee Appointment (153/168)} Procedures for selecting trustees and the term of office. & As per the governing document, four of the Trustee positions are appointed by virtue of their position within the Open University Students Association (OUSA). One further position is appointed by virtue of their previous position within OUSA. One Trustee is nominated by the Vice Chancellor of the Open University (OU)\noqa{spell-OU} and there are co-opted positions whereby the Trustees are empowered to approach up to two other persons to act as Trustees. It is envisaged that all Trustees will serve a~general term of two years in line with the main election periods within OUSA. \\\midrule
    \textsc{Reserves Policy (170/185)} What are the current financial reserves of the organization and how much these reserves should be as assumed? & The Trustees regularly reviews the amount of reserves that are required to ensure that they are adequate to fulfill the charities continuing obligations. \\\midrule
    \textsc{Income Summary (124/134)} General information on income for the last year, sometimes associated with information on expenses. & Excluding the adjustments for FRS17\noqa{spell-FRS17} in respect of Pension Fund the results by way of net incoming resources accumulated \noqa{spell-f3}f3.85m\noqa{spell-85m} as against \noqa{spell-E6}E6.78m\noqa{spell-78m} in 2014, however last years\noqa{grammar-POSSESSIVE_APOSTROPHE} performance benefited from extraordinary property sales generating a~profit of \noqa{grammar-UNIT_SPACE}\noqa{spell-15m}\noqa{spell-F3}F3.15m. \\\midrule
    \textsc{Auditor Opinion (190/192)} Summary of the opinion of an independent auditor or inspector, often in the form of a list of points. & In connection with my examination, no matter has come to my attention: 1. which gives me reasonable cause to believe that in any material respect the requirements to keep accounting records in accordance with Section 130 of the Charities Act; and to prepare accounts which accord with the accounting records and comply with the accounting requirements of the Charities Act have not been met; or 2. to which, in my opinion, attention should be drawn in order to enable a~proper understanding of the accounts to be reached. \\
    \bottomrule
    \end{tabular}
    \caption{Clauses annotated in Charity Annual Reports (one of three groups of documents included in the shared task). The values in parentheses indicate the number of documents with a~particular clause and the total number of clause instances, respectively. More examples are available in Appendix~C.\label{tab:clauses3}}
    \end{table*}
    \endgroup
}

\section{Competitive Baselines}\label{sec:method}

Solutions based on networks consuming pairs of sequences, such as BERT in sentence pair classification task setting~\cite{devlin2018bert}, are considered out of the scope of this paper since they are suboptimal in terms of performance--they require expensive encoding of all combinations from the Cartesian product between seeds and targets, making such solutions unsuitable for semantic similarity search due to the combinatorial explosion~\cite{reimers-2019-sentence-bert}. Because of the aforementioned problem and the fact that conventional classifiers require much more data than available in a~few-shot setting, in this section, we describe simple $k$-NN\noqa{spell-NN}-based approaches that we propose as baseline solutions to the problem stated.

\subsection{Processing Pipeline}

Evaluated solutions assume \noqa{spell-pre}pre-encoding of all candidate segments and can be described within the unified framework consisting of segmenters, vectorizers, projectors, aggregators, scorers, and choosers ordered in a pipeline of transformations.

\textit{Segmenter} is used to split a~text into candidate sub-sequences to be encoded and considered in further steps. All the described solutions rely on a~candidate sentence and n-grams of sentences, determined with the \textit{spaCy} CNN model trained on OntoNotes.\footnote{\url{http://github.com/explosion/spacy-models/releases/tag/en_core_web_sm-2.1.0}}
\textit{Vectorizer} produces vector representations of texts on either word, sub-word, or segment (e.g., ~sentence) level. In our case, vectorization was based on TF-IDF representations, static word embeddings, and neural sentence encoders.
\textit{Projector} projects embeddings into a~different space (e.g.,~decomposition methods such as PCA or ICA).
\textit{Aggregator} has the capability to use word or sub-word unit embeddings to create a~segment embedding (e.g.,~embedding mean, inverse frequency weighting, autoencoder).
\textit{Scorer} compares two or more embeddings and returns computed similarities. Since we often compare multiple seed embeddings with one embedding of a~candidate segment, a~scorer includes policies to aggregate scores obtained for multiple seeds into the final candidate score (e.g.,~mean of individual cosine similarities or max-pooling over Word Mover Distances).
\textit{Chooser} determines whether to return a~candidate segment with a~given score (e.g.,~threshold, one best per document, or a~combination thereof). For the sake of simplicity, during the evaluation, we restricted ourselves to the chooser returning only one, the most similar candidate. It is not optimal (because multiple might be expected), but we consider this setting a good reference for further methods. 

The proposed taxonomy is consistent with the assumptions made by~\citet{gillick2018endtoend}. It is presented in order to highlight the similarities and differences between particular solutions when they are introduced and compared within the ablation studies later in this paper. The next section describes vectorizers, aggregators, and scorers used for evaluation.

\subsubsection{Vectorizers}\label{sec:representations}

We intend to provide results of TF-IDF representations, as well as two methods that may be considered the state of the art of sentence embedding. The latter include \textit{Universal Sentence Encoder} (USE) and \textit{Sentence-BERT}.

USE is a~Trans\-former-based encoder, where an element-wise sum of word representations is treated as a sentence embedding~\cite{DBLP:journals/corr/abs-1803-11175}, trained with the multi-task objective. \textit{Sentence-BERT} is a modification of the pretrained BERT network, utilizing Siamese and triplet network structures to derive sentence embeddings, trained with the explicit objective of making them comparable with cosine similarity~\cite{reimers-2019-sentence-bert}. In both cases the original models released by the authors were used for the purposes of evaluation.

In addition, multiple contextual embeddings from Transformer-based language models, as well as static (context-less) GloVe word embeddings were tested~\cite{Pennington14glove:global}. Many approaches to generating context-dependent vector representations have been proposed in recent years (e.g.,~\citet{Peters:2018, DBLP:journals/corr/VaswaniSPUJGKP17}). One important advantage over static embeddings is the fact that every occurrence of the same word is assigned a~different embedding vector based on the context in which the word is used. Thus, it is much easier to address issues arising from pretrained static embeddings (e.g.,~taking into consideration polysemy of words). For the purposes of evaluation, we relied on Transformer-based models provided by authors of particular architectures, utilizing the Transformers library~\cite{Wolf2019HuggingFacesTS}. These include BERT~\cite{DBLP:journals/corr/abs-1810-04805}, GPT-1~\cite{Radford2018ImprovingLU}, GPT-2~\cite{gpt2}, and RoBERTa~\cite{liu2019roberta}. They differ substantially and introduce many innovations, though they are all based on either the encoder or the decoder from the original model proposed for sequence-to-sequence problems~\cite{DBLP:journals/corr/VaswaniSPUJGKP17}. Selected models were fine-tuned on using the next word prediction task on the Edgar corpus we release and re-evaluated.

\subsubsection{Aggregators}

In addition to conceptually simple methods such as average or max-polling operations, multiple solutions to utilizing word embeddings for comparing documents can be used. In addition to embeddings mean we evaluated the \textit{Smooth Inverse Frequency} (SIF), \textit{Word Mover's Distance} (WMD) and \textit{Discrete Cosine Transform} (DCT).

SIF is a~method proposed by~\citet{arora2017asimple}, where a representation of a~document is obtained in two steps. First, each word embedding is weighted by $a/(a + f_r)$, where $f_r$ stands for the underlying word's relative frequency, and $a$ is the weight parameter. Then, the projections on the first tSVD-calculated principal component are subtracted, providing final representations.

WMD is a~method of calculating a similarity between documents. For two documents, embeddings calculated for each word (e.g.,~with GloVe) are matched between documents, so that semantically similar pairs of words between documents are detected. This matching procedure generally leads to better results than simply averaging over embeddings for documents and calculating similarity between centers of mass of documents as their similarity~\cite{pmlr-v37-kusnerb15}. Recently, \citet{zhao2019moverscore} showed it might be beneficial to use the method with contextual word embeddings.

DCT is a~way to generate document-level representations in an order-preserving manner, adapted from image compression to NLP by~\citet{almarwani2019efficient}. After mapping an input sequence of real numbers to the coefficients of orthogonal cosine basis functions, low-order coefficients can be used as document embeddings, outperforming vector averaging on most tasks, as shown by the authors.

\subsection{Results}\label{sec:evaluation}

Table~\ref{tab:results} recapitulates the most important results of the completed evaluation.

\begin{table*}[t]
    \small\centering
    \begin{tabular}{lllllrrl}
        \toprule
        \textbf{Segmenter} &  \textbf{Vectorizer} &  \textbf{Projector} &  \textbf{Scorer} &  \textbf{Aggregator} & 
        \textbf{Soft} $\bm{F_1}$ \\
        \midrule
        sentence  & TF-IDF (1–2 grams, binary TF term) & — & mean cosine & — 
          & 0.38 \\
        & & tSVD (500)\footnote{TF-IDF with truncated SVD decomposition is commonly referred to as Latent Semantic Analysis~\cite{Halko:2011:FSR:2078879.2078881}.} & mean cosine & — 
          & \textbf{0.39} \\
        \midrule
        sentence  & GloVe (\noqa{spell-300d}300d, Wikipedia \& Gigaword) & — & mean cosine & mean 
          & 0.34 \\
          & & — & mean WMD & — 
          & 0.35  \\
        \vspace{8pt} & & SIF tSVD\footnote{SVD in SIF method is used to perform removal of single common component~\cite{arora2017asimple}.} & mean cosine & SIF 
          & 0.37 & \\
        sentence  & GloVe (\noqa{spell-300d}300d, EDGAR) & — & mean cosine & mean 
          & 0.36 \\
          & & — & mean WMD & — 
          & 0.35  \\
          & & SIF tSVD & mean cosine & SIF 
          & \textbf{0.41}  \\
        \midrule
        sentence  & Sentence-BERT (\noqa{spell-nli}\noqa{spell-stsb}base-nli-stsb-mean $\star$) & — & mean cosine  & mean 
        & 0.32  \\
        sentence  & USE (multilingual $\star$) & — & mean cosine  & — 
        & \textbf{0.38} \\ 
        \midrule
        sentence  & BERT, last layer (large-uncased-whole\ldots $\star$) & — & mean cosine & mean 
        & 0.35 \\
        sentence  & GPT-1, last layer & — & mean cosine & mean 
        & 0.36  \\
        sentence  & GPT-2, last layer (large $\star$) & — & mean cosine & mean 
        & 0.41 \\
        \vspace{8pt} sentence  & RoBERTa, last layer (large $\star$) & — & mean cosine & mean 
        & 0.31 \\
        sentence  & \noqa{grammar-ENGLISH_WORD_REPEAT_BEGINNING_RULE}GPT-1, last layer (fine-tuned) & — & mean cosine & mean 
        & 0.43  \\
        sentence & GPT-1, last layer (fine-tuned) & fICA (500) & mean cosine & mean 
        & 0.44 \\
        sentence  & GPT-2, last layer (large, fine-tuned) & — & mean cosine & mean 
        & 0.44  \\
        \vspace{8pt} sentence & GPT-2, last layer (large, fine-tuned) & fICA (400) & mean cosine & mean 
        & 0.45  \\
        1–3 sen. & GPT-1, last layer (fine-tuned) & & mean cosine & mean 
        & 0.47 \\
        1–3 sen. & GPT-1, last layer (fine-tuned) & fICA (500) & mean cosine & mean 
        & 0.49 \\
        1–3 sen. & GPT-2, last layer (large, fine-tuned) & & mean cosine & mean 
        & 0.46 \\
        1–3 sen. & GPT-2, last layer (large, fine-tuned) & fICA (400) & mean cosine & mean 
        & \textbf{0.51} \\
        \midrule
        human & & & & 
        & \textbf{0.84} \\
        \bottomrule
    \end{tabular}
    \caption{Selected results when returning a~single, most similar segment, determined with given segmenters, vectorizers, projectors, scorers and aggregators. The $\star$ symbol indicates only the best models from each architecture are presented here (results for the remaining ones are available in Appendix~B).}\label{tab:results}
\end{table*}

\begin{figure}
    \centering
    \includegraphics[width=0.42\textwidth]{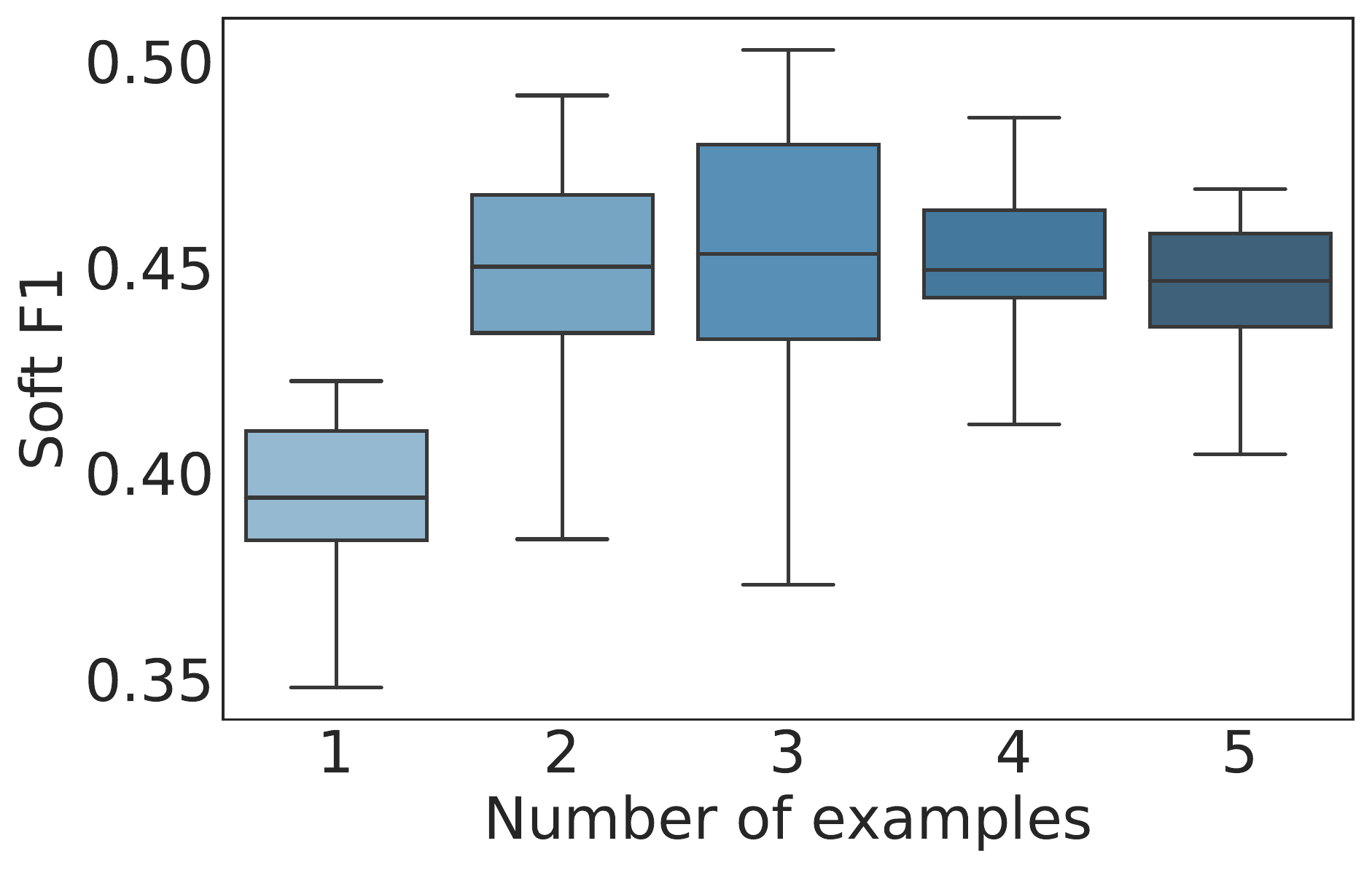}
    \caption{Performance as a~function of the number of example documents available (solutions based on LMs). The methods benefit substantially from availability of a~second example document and a~bigger number leads to a~decreased variance.
    }\label{fig:boxplot}
\end{figure}

Sentence-BERT and Universal Sentence Encoder could not outperform the simple TF-IDF approach, especially when SVD decomposition was applied (the setting commonly referred to as Latent Semantic Analysis). Static word embeddings with SIF weighting performed similarly to TF-IDF, or better, provided they were trained on a~legal text corpus rather than on general English. It could not be clearly confirmed whether the use of WMD or DCT is beneficial. For the latter, the best results were achieved with $c^0$, which in the case of the $k$-NN\noqa{spell-NN} algorithm leads to the same answers as mean-pooling and thus is not reported in the table. In case of $c^{0:n}$ where $n > 0$ constant decrease of $k$-NN methods performance was observed (Appendix~B).

Interestingly, from all the released USE models, the multilingual ones performed best — for the monolingual \textit{universal-sentence-encoder-large} model, scores were ten percentage points lower. The best Sentence-BERT model performed significantly \noqa{grammar-EXTREME_ADJECTIVES}worse than the best USE—note that the authors of Sentence-BERT compared it to monolingual models released earlier, which they indeed outperform. Moreover, Sentence-BERT does not perform better than BERT trained with whole word masking, although there is no Sentence-BERT equivalent of this model available so far.

In cases of averaging (sub)word embeddings from the last layer of neural Language Models, the results were either comparable or inferior to TF-IDF. The best-performing language models were GPT-1 and GPT-2. Fine-tuning of these on a~subsample of a~legal text corpus improved the results significantly, by a~factor of 3--7 points. LMs seem to benefit neither from SIF nor from the removal of a~single common component; their performance can, however, be mildly improved with a~conventionally used decomposition, such as ICA~\cite{Hyvrinen2000IndependentCA}.

Substantial improvement can be achieved by considering segments different from a~single sentence, such as n-grams of sentences (meaning that any contiguous sequence of up to $n$ sentences from a given text was scored and could be returned as a result).

Figure~\ref{fig:boxplot} presents how the performance of particular methods changes as a~function of the number of example documents available within the simple similarity averaging scheme used in all the presented solutions. In general, the methods benefit substantially from the availability of a~second example. A~bigger number leads to a~decreased variance but yields no improvement in the median score.

\section{Discussion}\label{sec:dicussion}

The brief evaluation presented in the previous section has multiple limitations. First, it assumed retrieval of a~single, most similar segment, whereas it appears that multiple clauses might be returned instead. However, we consider this restriction justifiable during a~preliminary comparison of applicable methods. Multiple alternative selectors may be proposed in the future.

Secondly, all the evaluated methods assume scoring with the policy of averaging individual similarities. We encourage readers to experiment with different pooling methods or meta-learning strategies. Moreover, even the LM-based methods we had studied the most can be further studied in the proposed shared task. For example, only embeddings from the last layer were evaluated, even though it is possible that the higher layers may capture semantics better.

Finally, it is in principle possible to address the task in entirely different ways, for example, by performing neither segmentation nor aggregation of word embeddings at all, but by matching clauses on the word level instead, which may be an interesting direction for further research. We decided to take the most common and straightforward way, due to fact performed evaluations are to serve as baselines for other methods. 

\section{Related Work}\label{sec:related-work}

There is a~large and varied body of work related to information retrieval in general; however, following~\citet{gillick2018endtoend} we consider the problem stated in an end-to-end \noqa{grammar-IN_A_X_MANNER}manner, where the nearest neighbor search is performed on dense document representations. With this assumption, the main issue is to obtain reliable representations of documents, \noqa{grammar-WHERE_AS}where by document we mean \textit{any self-contained unit that can be returned to the user as a~search result}~\cite{books/daglib/0031897}. We use the term \textit{segment} with the same meaning wherever it aids clarity.

Many approaches considered in the literature rely on word embedding and aggregation strategies. Simple methods proposed include averaging, as in the continuous bag-of-words (CBOW) model~\cite{41224} or frequency-weighted averaging with the decomposition method applied~\cite{arora2017asimple}. More sophisticated schemes include utilizing multiple weights, such as a~novelty score, a~significance score, and a~corpus-wise uniqueness~\cite{Yang2018ZerotrainingSE} or computing a~vector of locally aggregated descriptors~\cite{ionescu2019vector}. Most of the proposed methods are orderless, and their limitations were recently discussed by~\citet{DBLP:journals/corr/abs-1902-06423}. However, there are also pooling approaches preserving spatial information, such as a~hierarchical pooling operation~\cite{shen2018baseline}. Other methods of obtaining sentence representations from word embeddings include training an autoencoder on a~large collection of unlabeled data~\cite{zhang2018learning} or utilizing random encoders~\cite{wieting2019training}. Despite its shortcomings and the availability of many sophisticated alternatives, the CBOW model is a~common choice due to its ability to ensure strong results on many downstream tasks.

Different approaches assume training encoders through document embedding in an~unsupervised or supervised manner, without the need for explicit aggregation. The former include Skip-Thought Vectors, trained with the objective of reconstructing the surrounding sentences of an encoded passage~\cite{NIPS2015_5950}. Although this method was outperformed by supervised models trained on a~single NLI task~\cite{conneau2017supervised}, paraphrase corpora~\cite{jiao-etal-2018-convolutional} or multiple tasks~\cite{DBLP:journals/corr/abs-1804-00079}, the objective of predicting the next sentence is used as an additional objective in multiple novel models, such as the Universal Sentence Encoder~\cite{DBLP:journals/corr/abs-1803-11175}. Even though many Transformer-based language models implement their own pooling strategy for generating sentence representations (special token pooling), they were shown to yield weak sentence embeddings, as described recently by~\citet{reimers-2019-sentence-bert}. The authors proposed a~superior method of fine-tuning a~pretrained BERT network with Siamese and triplet network structures to obtain sentence embeddings.

There were attempts to utilize semantic similarity methods explicitly in the legal domain, e.g.,~for a~case law entailment within the COLIEE shared task. In a~recent edition, \citet{Rabelo:2019:CST:3322640.3326741} used a~BERT model fine-tuned on a~provided training set in a~supervised manner, and achieved the highest F-score among all teams. However, due to the reasons discussed in Section~\ref{sec:method}, their approach is not consistent with the nearest neighbor search, which is what we are aiming for.


\section{Summary and Conclusions}\label{sec:summary}

We have introduced a~new shared task of semantic retrieval from legal texts, which differs substantially from conventional NLI. It is heavily inspired by enterprise solutions referred to as \textit{contract discovery}, focused on ensuring the inclusion of relevant clauses or their retrieval for further analysis. The main distinguishing characteristic of Contract Discovery shared task is conceptual, since:
\begin{itemize}
    \item Candidate sequences are being mined from real texts. It is assumed span identification should be performed (systems should be able to return any document substring without any segmentation given in advance).
    \item It is suited for few-shot methods, filling the gap between conventional sentence classification and NLI tasks based on sentence pairs.
\end{itemize}

For the purposes of providing competetive baselines, we considered the problem stated in an end-to-end \noqa{grammar-IN_A_X_MANNER}manner, where the nearest neighbor search is performed on document representations. With this assumption, the main issue was to obtain representations of text fragments, which we referred to as segments. The description of the task was followed by the evaluation of multiple $k$-NN\noqa{spell-NN}-based solutions within the unified framework, which may be used to describe future solutions. Moreover, a~practical justification for handling the problem with $k$-NN was briefly introduced.

It has been shown that in this particular setting, pretrained, \textit{universal} encoders fail to provide satisfactory results. One may suspect that this is a~result of the difference between the domain they were trained on and the legal domain. During the evaluation, solutions based on the Language Models performed well, especially when unsupervised fine-tuning was applied. In addition to the aforementioned ability to fine-tune the method on legal texts, the most important indicator of success so far has been the involvement of multiple, sometimes overlapping substrings instead of sentences. Moreover, it has been demonstrated that the methods benefit substantially from the availability of a~second example, and the presence of more leads to a~decrease in variance, even when a~simple similarity averaging scheme is applied.

The discussion regarding the presented methods and their limitations briefly outlined possible measures towards improving the baseline methods. In addition to the dataset and reference results, legal-specialized LMs have been made released to assist the research community in performing further experiments.

The Contract Discovery dataset, Edgar Corpus, we crawled, and all the mentioned models are publicly available on GitHub: \url{https://github.com/applicaai/contract-discovery}.

\section*{Acknowledgements}

The Smart Growth Operational Programme supported this research under project no. POIR.01.01.01-00-0605/19 (\textit{Disruptive adoption of Neural Language Modelling for automation of text-intensive work}).

There are no copyright issues regarding the Contract Discovery dataset, as both sources belong to the public domain. Documents were annotated ethically by our co-workers. Moreover, the colleagues who participated in annotation are among the authors of the paper.

\bibliographystyle{acl_natbib}
\bibliography{bibliography}

\begin{thebibliography}{50}
\expandafter\ifx\csname natexlab\endcsname\relax\def\natexlab#1{#1}\fi

\bibitem[{Almarwani et~al.(2019)Almarwani, Aldarmaki, and
  Diab}]{almarwani2019efficient}
Nada Almarwani, Hanan Aldarmaki, and Mona Diab. 2019.
\newblock \href {http://arxiv.org/abs/1909.03104} {Efficient sentence embedding
  using discrete cosine transform}.

\bibitem[{Arora et~al.(2017)Arora, Liang, and Ma}]{arora2017asimple}
Sanjeev Arora, Yingyu Liang, and Tengyu Ma. 2017.
\newblock A simple but tough-to-beat baseline for sentence embeddings.

\bibitem[{Bao et~al.(2019)Bao, Wu, Chang, and Barzilay}]{bao2019few}
Yujia Bao, Menghua Wu, Shiyu Chang, and Regina Barzilay. 2019.
\newblock Few-shot text classification with distributional signatures.
\newblock \emph{arXiv:1908.06039}.

\bibitem[{Baron et~al.(2006)Baron, Archives, Administration, and
  General}]{Baron06trec-2006legal}
Jason~R. Baron, National Archives, Records Administration, and Office~Of
  General. 2006.
\newblock Trec-2006 legal track overview.
\newblock In \emph{In The Fifteenth Text REtrieval Conference (TREC 2006)
  Proceedings}.

\bibitem[{Bowman et~al.(2015)Bowman, Angeli, Potts, and
  Manning}]{snli:emnlp2015}
Samuel~R. Bowman, Gabor Angeli, Christopher Potts, and Christopher~D. Manning.
  2015.
\newblock A large annotated corpus for learning natural language inference.
\newblock In \emph{Proceedings of the 2015 Conference on Empirical Methods in
  Natural Language Processing (EMNLP)}. Association for Computational
  Linguistics.

\bibitem[{Burkard et~al.(2012)Burkard, Dell'Amico, and
  Martello}]{c225e96c4acd499eb7a5ceabd178a543}
Rainer Burkard, Mauro Dell'Amico, and Silvano Martello. 2012.
\newblock \emph{Assignment Problems. Revised reprint.}
\newblock SIAM - Society of Industrial and Applied Mathematics.
\newblock 393 Seiten.

\bibitem[{Büttcher et~al.(2010)Büttcher, Clarke, and
  Cormack}]{books/daglib/0031897}
Stefan Büttcher, Charles L.~A. Clarke, and Gordon~V. Cormack. 2010.
\newblock \emph{Information Retrieval - Implementing and Evaluating Search
  Engines.}
\newblock MIT Press.

\bibitem[{Cer et~al.(2018)Cer, Yang, Kong, Hua, Limtiaco, John, Constant,
  Guajardo{-}Cespedes, Yuan, Tar, Sung, Strope, and
  Kurzweil}]{DBLP:journals/corr/abs-1803-11175}
Daniel Cer, Yinfei Yang, Sheng{-}yi Kong, Nan Hua, Nicole Limtiaco, Rhomni~St.
  John, Noah Constant, Mario Guajardo{-}Cespedes, Steve Yuan, Chris Tar,
  Yun{-}Hsuan Sung, Brian Strope, and Ray Kurzweil. 2018.
\newblock \href {http://arxiv.org/abs/1803.11175} {Universal sentence encoder}.
\newblock \emph{CoRR}, abs/1803.11175.

\bibitem[{Chu(2011)}]{Chu2011FactorsAR}
Heting Chu. 2011.
\newblock Factors affecting relevance judgment: a report from trec legal track.
\newblock \emph{Journal of Documentation}, 67:264--278.

\bibitem[{Conneau et~al.(2017)Conneau, Kiela, Schwenk, Barrault, and
  Bordes}]{conneau2017supervised}
Alexis Conneau, Douwe Kiela, Holger Schwenk, Loic Barrault, and Antoine Bordes.
  2017.
\newblock \href {http://arxiv.org/abs/1705.02364} {Supervised learning of
  universal sentence representations from natural language inference data}.

\bibitem[{Da~San~Martino et~al.(2020)Da~San~Martino, Barr\'{o}n-Cede\~no,
  Wachsmuth, Petrov, and Nakov}]{DaSanMartinoSemeval20task11}
Giovanni Da~San~Martino, Alberto Barr\'{o}n-Cede\~no, Henning Wachsmuth,
  Rostislav Petrov, and Preslav Nakov. 2020.
\newblock {SemEval}-2020 task 11: Detection of propaganda techniques in news
  articles.
\newblock In \emph{Proceedings of the 14th International Workshop on Semantic
  Evaluation}, SemEval 2020, Barcelona, Spain.

\bibitem[{Devlin et~al.(2018{\natexlab{a}})Devlin, Chang, Lee, and
  Toutanova}]{devlin2018bert}
Jacob Devlin, Ming-Wei Chang, Kenton Lee, and Kristina Toutanova.
  2018{\natexlab{a}}.
\newblock {BERT}: Pre-training of deep bidirectional transformers for language
  understanding.
\newblock \emph{arXiv preprint arXiv:1810.04805}.

\bibitem[{Devlin et~al.(2018{\natexlab{b}})Devlin, Chang, Lee, and
  Toutanova}]{DBLP:journals/corr/abs-1810-04805}
Jacob Devlin, Ming{-}Wei Chang, Kenton Lee, and Kristina Toutanova.
  2018{\natexlab{b}}.
\newblock \href {http://arxiv.org/abs/1810.04805} {{BERT:} pre-training of deep
  bidirectional transformers for language understanding}.
\newblock \emph{CoRR}, abs/1810.04805.

\bibitem[{Fabian Caba~Heilbron and Niebles(2015)}]{caba2015activitynet}
Bernard~Ghanem Fabian Caba~Heilbron, Victor~Escorcia and Juan~Carlos Niebles.
  2015.
\newblock Activitynet: A large-scale video benchmark for human activity
  understanding.
\newblock In \emph{Proceedings of the IEEE Conference on Computer Vision and
  Pattern Recognition}, pages 961--970.

\bibitem[{Fritzler et~al.(2019)Fritzler, Logacheva, and
  Kretov}]{Fritzler:2019:FCN:3297280.3297378}
Alexander Fritzler, Varvara Logacheva, and Maksim Kretov. 2019.
\newblock \href {https://doi.org/10.1145/3297280.3297378} {Few-shot
  classification in named entity recognition task}.
\newblock In \emph{Proceedings of the 34th ACM/SIGAPP Symposium on Applied
  Computing}, SAC '19, pages 993--1000, New York, NY, USA. ACM.

\bibitem[{Gillick et~al.(2018)Gillick, Presta, and Tomar}]{gillick2018endtoend}
Daniel Gillick, Alessandro Presta, and Gaurav~Singh Tomar. 2018.
\newblock \href {http://arxiv.org/abs/1811.08008} {End-to-end retrieval in
  continuous space}.

\bibitem[{Grali{\'n}ski et~al.(2019)Grali{\'n}ski, Wr{\'o}blewska,
  Stani\-s{\l}awek, Grabowski, and G{\'o}recki}]{gralinski-etal-2019-geval}
Filip Grali{\'n}ski, Anna Wr{\'o}blewska, Tomasz Stani\-s{\l}awek, Kamil
  Grabowski, and Tomasz G{\'o}recki. 2019.
\newblock \href {https://www.aclweb.org/anthology/W19-4826} {{GE}val: Tool for
  debugging {NLP} datasets and models}.
\newblock In \emph{Proceedings of the 2019 ACL Workshop BlackboxNLP: Analyzing
  and Interpreting Neural Networks for NLP}, pages 254--262, Florence, Italy.
  Association for Computational Linguistics.

\bibitem[{Halko et~al.(2011)Halko, Martinsson, and
  Tropp}]{Halko:2011:FSR:2078879.2078881}
N.~Halko, P.~G. Martinsson, and J.~A. Tropp. 2011.
\newblock \href {https://doi.org/10.1137/090771806} {Finding structure with
  randomness: Probabilistic algorithms for constructing approximate matrix
  decompositions}.
\newblock \emph{SIAM Rev.}, 53(2):217--288.

\bibitem[{Hyv{\"a}rinen and Oja(2000)}]{Hyvrinen2000IndependentCA}
Aapo Hyv{\"a}rinen and Erkki Oja. 2000.
\newblock Independent component analysis: algorithms and applications.
\newblock \emph{Neural networks : the official journal of the International
  Neural Network Society}, 13 4-5:411--30.

\bibitem[{Ionescu and Butnaru(2019)}]{ionescu2019vector}
Radu~Tudor Ionescu and Andrei~M. Butnaru. 2019.
\newblock \href {http://arxiv.org/abs/1902.08850} {{Vector of
  Locally-Aggregated Word Embeddings (VLAWE): A Novel Document-level
  Representation}}.

\bibitem[{Jiang et~al.(2014)Jiang, Liu, Roshan~Zamir, Toderici, Laptev, Shah,
  and Sukthankar}]{THUMOS14}
Y.-G. Jiang, J.~Liu, A.~Roshan~Zamir, G.~Toderici, I.~Laptev, M.~Shah, and
  R.~Sukthankar. 2014.
\newblock {THUMOS} challenge: Action recognition with a large number of
  classes.
\newblock \url{http://crcv.ucf.edu/THUMOS14/}.

\bibitem[{Jiao et~al.(2018)Jiao, Wang, and Feng}]{jiao-etal-2018-convolutional}
Xiaoqi Jiao, Fang Wang, and Dan Feng. 2018.
\newblock \href {https://www.aclweb.org/anthology/C18-1209} {Convolutional
  neural network for universal sentence embeddings}.
\newblock In \emph{Proceedings of the 27th International Conference on
  Computational Linguistics}, pages 2470--2481, Santa Fe, New Mexico, USA.
  Association for Computational Linguistics.

\bibitem[{Kano et~al.(2017)Kano, Kim, Goebel, and Satoh}]{Kano2017OverviewOC}
Yoshinobu Kano, Mi~Young Kim, Randy Goebel, and Ken Satoh. 2017.
\newblock {Overview of COLIEE 2017}.
\newblock In \emph{COLIEE@ICAIL}.

\bibitem[{Kiros et~al.(2015)Kiros, Zhu, Salakhutdinov, Zemel, Urtasun,
  Torralba, and Fidler}]{NIPS2015_5950}
Ryan Kiros, Yukun Zhu, Ruslan~R Salakhutdinov, Richard Zemel, Raquel Urtasun,
  Antonio Torralba, and Sanja Fidler. 2015.
\newblock \href {http://papers.nips.cc/paper/5950-skip-thought-vectors.pdf}
  {Skip-thought vectors}.
\newblock In C.~Cortes, N.~D. Lawrence, D.~D. Lee, M.~Sugiyama, and R.~Garnett,
  editors, \emph{Advances in Neural Information Processing Systems 28}, pages
  3294--3302. Curran Associates, Inc.

\bibitem[{Kusner et~al.(2015)Kusner, Sun, Kolkin, and
  Weinberger}]{pmlr-v37-kusnerb15}
Matt Kusner, Yu~Sun, Nicholas Kolkin, and Kilian Weinberger. 2015.
\newblock \href {http://proceedings.mlr.press/v37/kusnerb15.html} {From word
  embeddings to document distances}.
\newblock In \emph{Proceedings of the 32nd International Conference on Machine
  Learning}, volume~37 of \emph{Proceedings of Machine Learning Research},
  pages 957--966, Lille, France. PMLR.

\bibitem[{Liu et~al.(2019)Liu, Ott, Goyal, Du, Joshi, Chen, Levy, Lewis,
  Zettlemoyer, and Stoyanov}]{liu2019roberta}
Yinhan Liu, Myle Ott, Naman Goyal, Jingfei Du, Mandar Joshi, Danqi Chen, Omer
  Levy, Mike Lewis, Luke Zettlemoyer, and Veselin Stoyanov. 2019.
\newblock \href {http://arxiv.org/abs/1907.11692} {{RoBERTa: A Robustly
  Optimized BERT Pretraining Approach}}.

\bibitem[{Mai et~al.(2019)Mai, Galke, and
  Scherp}]{DBLP:journals/corr/abs-1902-06423}
Florian Mai, Lukas Galke, and Ansgar Scherp. 2019.
\newblock \href {http://arxiv.org/abs/1902.06423} {{CBOW} is not all you need:
  Combining {CBOW} with the compositional matrix space model}.
\newblock \emph{CoRR}, abs/1902.06423.

\bibitem[{Maxwell and Schafer(2008)}]{Maxwell2008ConceptAC}
K.~Tamsin Maxwell and Burkhard Schafer. 2008.
\newblock Concept and context in legal information retrieval.
\newblock In \emph{JURIX}.

\bibitem[{Mikolov et~al.(2013)Mikolov, Chen, Corrado, and Dean}]{41224}
Tomas Mikolov, Kai Chen, Greg~S. Corrado, and Jeffrey Dean. 2013.
\newblock \href {http://arxiv.org/abs/1301.3781} {Efficient estimation of word
  representations in vector space}.

\bibitem[{Nagpal et~al.(2018)Nagpal, Wadhwa, Gupta, Shaikh, Mehta, and
  Goyal}]{DBLP:journals/corr/abs-1809-04262}
Rashmi Nagpal, Chetna Wadhwa, Mallika Gupta, Samiulla Shaikh, Sameep Mehta, and
  Vikram Goyal. 2018.
\newblock \href {http://arxiv.org/abs/1809.04262} {Extracting fairness policies
  from legal documents}.
\newblock \emph{CoRR}, abs/1809.04262.

\bibitem[{Oard et~al.(2010)Oard, Baron, Hedin, Lewis, and
  Tomlinson}]{oard2010evaluation}
W.~Douglas Oard, R.~Jason Baron, Bruce Hedin, D.~David Lewis, and Stephen
  Tomlinson. 2010.
\newblock Evaluation of information retrieval for e-discovery.
\newblock \emph{Artif. Intell. Law}, pages 347--386.

\bibitem[{Pennington et~al.(2014)Pennington, Socher, and
  Manning}]{Pennington14glove:global}
Jeffrey Pennington, Richard Socher, and Christopher~D. Manning. 2014.
\newblock {GloVe: Global vectors for word representation}.
\newblock In \emph{In EMNLP}.

\bibitem[{Peters et~al.(2018)Peters, Neumann, Iyyer, Gardner, Clark, Lee, and
  Zettlemoyer}]{Peters:2018}
Matthew~E. Peters, Mark Neumann, Mohit Iyyer, Matt Gardner, Christopher Clark,
  Kenton Lee, and Luke Zettlemoyer. 2018.
\newblock Deep contextualized word representations.
\newblock In \emph{Proc. of NAACL}.

\bibitem[{Potthast et~al.(2010)Potthast, Stein, Barr{\'o}n-Cede{\~n}o, and
  Rosso}]{potthast-etal-2010-evaluation}
Martin Potthast, Benno Stein, Alberto Barr{\'o}n-Cede{\~n}o, and Paolo Rosso.
  2010.
\newblock \href {https://www.aclweb.org/anthology/C10-2115} {An evaluation
  framework for plagiarism detection}.
\newblock In \emph{Coling 2010: Posters}, pages 997--1005, Beijing, China.
  Coling 2010 Organizing Committee.

\bibitem[{Rabelo et~al.(2019)Rabelo, Kim, and
  Goebel}]{Rabelo:2019:CST:3322640.3326741}
Juliano Rabelo, Mi-Young Kim, and Randy Goebel. 2019.
\newblock \href {https://doi.org/10.1145/3322640.3326741} {Combining similarity
  and transformer methods for case law entailment}.
\newblock In \emph{Proceedings of the Seventeenth International Conference on
  Artificial Intelligence and Law}, ICAIL '19, pages 290--296, New York, NY,
  USA. ACM.

\bibitem[{Radford(2018)}]{Radford2018ImprovingLU}
Alec Radford. 2018.
\newblock Improving language understanding by generative pre-training.

\bibitem[{Radford et~al.(2018)Radford, Wu, Child, Luan, Amodei, and
  Sutskever}]{gpt2}
Alec Radford, Jeffrey Wu, Rewon Child, David Luan, Dario Amodei, and Ilya
  Sutskever. 2018.
\newblock \href
  {https://d4mucfpksywv.cloudfront.net/better-language-models/language-models.pdf}
  {Language models are unsupervised multitask learners}.

\bibitem[{Reimers and Gurevych(2019)}]{reimers-2019-sentence-bert}
Nils Reimers and Iryna Gurevych. 2019.
\newblock \href {http://arxiv.org/abs/1908.10084} {{Sentence-BERT: Sentence
  Embeddings using Siamese BERT-Networks}}.
\newblock In \emph{Proceedings of the 2019 Conference on Empirical Methods in
  Natural Language Processing}. Association for Computational Linguistics.

\bibitem[{Shen et~al.(2018)Shen, Wang, Wang, Min, Su, Zhang, Li, Henao, and
  Carin}]{shen2018baseline}
Dinghan Shen, Guoyin Wang, Wenlin Wang, Martin~Renqiang Min, Qinliang Su, Yizhe
  Zhang, Chunyuan Li, Ricardo Henao, and Lawrence Carin. 2018.
\newblock \href {http://arxiv.org/abs/1805.09843} {Baseline needs more love: On
  simple word-embedding-based models and associated pooling mechanisms}.

\bibitem[{Subramanian et~al.(2018)Subramanian, Trischler, Bengio, and
  Pal}]{DBLP:journals/corr/abs-1804-00079}
Sandeep Subramanian, Adam Trischler, Yoshua Bengio, and Christopher~J. Pal.
  2018.
\newblock \href {http://arxiv.org/abs/1804.00079} {Learning general purpose
  distributed sentence representations via large scale multi-task learning}.
\newblock \emph{CoRR}, abs/1804.00079.

\bibitem[{Tejedor et~al.(2019)Tejedor, Toledano, Lopez-Otero, Docio-Fernandez,
  Pe\~{n}agarikano, Rodriguez-Fuentes, and
  Moreno-Sandoval}]{10.1186/s13636-019-0156-x}
Javier Tejedor, Doroteo~T. Toledano, Paula Lopez-Otero, Laura Docio-Fernandez,
  Mikel Pe\~{n}agarikano, Luis~Javier Rodriguez-Fuentes, and Antonio
  Moreno-Sandoval. 2019.
\newblock \href {https://doi.org/10.1186/s13636-019-0156-x} {{Search on Speech
  from Spoken Queries: The Multi-Domain International ALBAYZIN 2018
  Query-by-Example Spoken Term Detection Evaluation}}.
\newblock \emph{EURASIP J. Audio Speech Music Process.}, 2019(1).

\bibitem[{Vanderbeck et~al.(2011)Vanderbeck, Bockhorst, and
  Oldfather}]{Vanderbeck2011AML}
Scott Vanderbeck, Joseph Bockhorst, and Chad Oldfather. 2011.
\newblock A machine learning approach to identifying sections in legal briefs.
\newblock In \emph{MAICS}.

\bibitem[{Vaswani et~al.(2017)Vaswani, Shazeer, Parmar, Uszkoreit, Jones,
  Gomez, Kaiser, and Polosukhin}]{DBLP:journals/corr/VaswaniSPUJGKP17}
Ashish Vaswani, Noam Shazeer, Niki Parmar, Jakob Uszkoreit, Llion Jones,
  Aidan~N. Gomez, Lukasz Kaiser, and Illia Polosukhin. 2017.
\newblock \href {http://arxiv.org/abs/1706.03762} {Attention is all you need}.
\newblock \emph{CoRR}, abs/1706.03762.

\bibitem[{Wang et~al.(2019)Wang, Yao, Kwok, and Ni}]{Wang2019GeneralizingFA}
Yaqing Wang, Quanming Yao, James Kwok, and Lionel~M. Ni. 2019.
\newblock Generalizing from a few examples: A survey on few-shot learning.

\bibitem[{Wieting and Kiela(2019)}]{wieting2019training}
John Wieting and Douwe Kiela. 2019.
\newblock \href {http://arxiv.org/abs/1901.10444} {No training required:
  Exploring random encoders for sentence classification}.

\bibitem[{Williams et~al.(2017)Williams, Nangia, and Bowman}]{Williams2017ABC}
Adina Williams, Nikita Nangia, and Samuel~R. Bowman. 2017.
\newblock A broad-coverage challenge corpus for sentence understanding through
  inference.
\newblock In \emph{NAACL-HLT}.

\bibitem[{Wolf et~al.(2019)Wolf, Debut, Sanh, Chaumond, Delangue, Moi, Cistac,
  Rault, Louf, Funtowicz, and Brew}]{Wolf2019HuggingFacesTS}
Thomas Wolf, Lysandre Debut, Victor Sanh, Julien Chaumond, Clement Delangue,
  Anthony Moi, Pierric Cistac, Tim Rault, R'emi Louf, Morgan Funtowicz, and
  Jamie Brew. 2019.
\newblock {HuggingFace's Transformers: State-of-the-art Natural Language
  Processing}.
\newblock \emph{ArXiv}, abs/1910.03771.

\bibitem[{Yang et~al.(2018)Yang, Zhu, and Chen}]{Yang2018ZerotrainingSE}
Ziyi Yang, Chenguang Zhu, and Weizhu Chen. 2018.
\newblock Zero-training sentence embedding via orthogonal basis.
\newblock \emph{ArXiv}, abs/1810.00438.

\bibitem[{Zhang et~al.(2018)Zhang, Wu, Li, and Li}]{zhang2018learning}
Minghua Zhang, Yunfang Wu, Weikang Li, and Wei Li. 2018.
\newblock \href {http://arxiv.org/abs/1809.06590} {Learning universal sentence
  representations with mean-max attention autoencoder}.

\bibitem[{Zhao et~al.(2019)Zhao, Peyrard, Liu, Gao, Meyer, and
  Eger}]{zhao2019moverscore}
Wei Zhao, Maxime Peyrard, Fei Liu, Yang Gao, Christian~M. Meyer, and Steffen
  Eger. 2019.
\newblock {MoverScore: Text Generation Evaluating with Contextualized
  Embeddings and Earth Mover Distance}.
\newblock In \emph{Proceedings of the 2019 Conference on Empirical Methods in
  Natural Language Processing}, Hong Kong, China. Association for Computational
  Linguistics.

\end{thebibliography}

\appendix

\appendix


\section{File Structure}
The documents’ content can be found in the \code{reference.tsv} files. The input files \code{in.tsv} consist of tab-separated fields: Target ID (e.g.~\textit{57}), Clause considered (e.g.~\textit{governing-law}), Example \#1 (e.g.~\textit{59 15215-15453}\noqa{latex-8}), \ldots, Example \#N. Each example consists of document ID and characters range. Ranges can be discontinuous. In such a~case the sequences are separated with a~comma, e.g.~4103-4882,12127-12971. The file with answers (\code{expected.tsv}) contains one answer per line, consisting of the entity name (to be copied from input) and characters range in the same format as described above. The reference file contains two tab-separated fields: document ID and content.

\section{Other Evaluation Results}

Tables below describe evaluation results which were not included in the paper (or were included without broader context, that is without reference to different results from the same class of solutions).

Table~\ref{sbert} presents results with all the evaluated Sentence-BERT models. Table~\ref{tf} shows scores achieved by TF-IDF with different settings, including other n-gram ranges. Results of particular Universal Sentence Encoder models are presented in Table~\ref{use}. Table~\ref{trans} shows results of Transformer-based Language Models not included in the paper. Finally, Table~\ref{dct} is devoted to analysis of Discrete Cosine Transform embeddings.

\begin{table}[!htbp]
\footnotesize
\centering
\begin{tabular}{p{5.5cm} r}
        \toprule
        \textbf{Model} & \textbf{Soft} $\bm{F_1}$  \\
        \midrule
        bert-base-nli-cls-token & 0.29 \\
        bert-base-nli-max-tokens & 0.30 \\
        bert-base-nli-mean-tokens & 0.31 \\
        bert-base-nli-stsb-mean-tokens & \textbf{0.32} \\
        bert-base-wikipedia-sections-mean-tokens & 0.25 \\
        bert-large-nli-cls-token & 0.29 \\
        bert-large-nli-max-tokens & 0.30 \\
        bert-large-nli-mean-tokens & 0.30 \\
        \vspace{3pt} bert-large-nli-stsb-mean-tokens & 0.31 \\
        roberta-base-nli-mean-tokens & 0.28 \\
        roberta-base-nli-stsb-mean-tokens & 0.29 \\
        roberta-large-nli-mean-tokens & 0.31 \\
        roberta-large-nli-stsb-mean-tokens & 0.31 \\
        \bottomrule
\end{tabular}
\caption{Results of Sentence-BERT models on the \textit{test-A} dataset when returning the most similar sentence. Names as in \textit{sentence-transformers} library: \url{https://github.com/UKPLab/sentence-transformers}\label{sbert}}
\end{table}

\begin{table}[!htbp]
\footnotesize
\centering
\begin{tabular}{p{4cm} p{1cm} r}
      \toprule
      \textbf{Range (n-grams)} & \textbf{Binary} & \textbf{Soft} $\bm{F_1}$  \\
      \midrule
     1–1 & $-$ & 0.32 \\
     1–2 & $-$ & 0.35 \\
     1–3 & $-$ & 0.36 \\
     1–1 & $+$ & 0.36 \\
     1–2 & $+$ & \textbf{0.38} \\
     1–3 & $+$ & 0.37 \\
     \bottomrule
     \end{tabular}
     \caption{Results of TF-IDF on the \textit{test-A} dataset when returning the most similar sentence.\label{tf}}
\end{table}

\begin{table}[!htbp]
\footnotesize
\centering
\begin{tabular}{p{5.5cm} r}
      \toprule
      \textbf{Model} & \textbf{Soft} $\bm{F_1}$  \\
      \midrule
      multilingual/1 & \textbf{0.38} \\
      multilingual-large/1 & 0.33 \\
      multilingual-qa/1 & 0.28 \\
      large/3 & 0.26 \\
     \bottomrule
     \end{tabular}
     \caption{Results of Universal Sentence Encoder models on the \textit{test-A} dataset when returning the most similar sentence.\label{use}}
\end{table}

\begin{table}[!htbp]
\footnotesize
\centering
\begin{tabular}{p{5.5cm} r}
      \toprule
      \textbf{Model} & \textbf{Soft} $\bm{F_1}$  \\
      \midrule
      bert-base-cased & 0.25 \\
      bert-base-multilingual-cased & 0.24 \\
      bert-base-multilingual-uncased & 0.32 \\
      bert-base-uncased & 0.26 \\
      bert-large-cased & 0.21 \\
      bert-large-cased-whole-word-masking & 0.31 \\
      bert-large-uncased & 0.18 \\
      \vspace{3pt} bert-large-uncased-whole-word-masking & 0.35 \\
      roberta-base & 0.25 \\
      \vspace{3pt} roberta-large & 0.32 \\
      \vspace{3pt} openai-gpt & 0.36 \\
      gpt2 & 0.16 \\
      gpt2-medium & 0.11 \\
      gpt2-large & \textbf{0.41} \\
     \bottomrule
     \end{tabular}
     \caption{Results of particular Transformer-based Language Models (without finetuning) on the \textit{test-A} dataset when returning the most similar sentence. Names as in \textit{transformers} library: \url{https://github.com/huggingface/transformers}\label{trans}}
\end{table}

\begin{table}[!htbp]
\footnotesize
\centering
\begin{tabular}{p{5.5cm} r}
     \toprule
     \textbf{C} & \textbf{Soft} $\bm{F_1}$  \\
     \midrule
     $c^0$ & \textbf{0.36} \\
     $c^{0:1}$ & 0.30 \\
     $c^{0:2}$ & 0.25 \\
     $c^{0:3}$ & 0.20 \\
     $c^{0:4}$ & 0.18 \\
     \bottomrule
     \end{tabular}
     \caption{Results of GloVe embeddings (300d, EDGAR) on the \textit{test-A} dataset when Discrete Cosine Transform sentence embeddings were created. The $c^0$ is equivalent to embeddings mean when $k$-NN methods are considered. The similar decrease of performance was observed for other models.\label{dct}}
\end{table}

\onecolumn

\section{Rest of the Clauses Considered}

Random subsets of bond issue prospectuses and non-disclosure agreement documents from the US EDGAR database\footnote{\url{http://www.www.sec.gov/edgar.shtml}}, as well as annual reports of charitable organizations from the UK Charity Register\footnote{\url{http://www.gov.uk/find-charity-information}} were annotated, in such a~way that clauses of the same type were selected (e.g.~determining the governing law, merger restrictions, tax changes call or reserves policy). Clause types depend on the type of a legal act and can consist of a~single sentence, multiple sentences or sentence fragments. Tables bellow present clause types annotated in each of the document groups.

{\small
    \begingroup
    \begin{longtable}{p{0.37\textwidth} p{0.57\textwidth}}
    \toprule
    \textbf{Clause (Instances)} & \textbf{Example} \\
    \midrule
    \textsc{Governing Law (152/160)} The parties agree on which jurisdiction the contract will be subject to. & This Agreement shall be governed by and construed in accordance with the laws of the State of California without reference to its rules of conflicts of laws. \\\midrule
    \textsc{Confidential Period (108/122)} The parties undertake to maintain confidentiality for a certain period of time. & The term of this Agreement during which Confidential Information may be disclosed by one Party to the other Party shall begin on the Effective Date and end five (5) years after the Effective Date, unless extended by mutual agreement. \\\midrule
    \textsc{Effective Date (79/89)} Information on the date of entry into force of the contract. & THIS AGREEMENT is entered into as of the 30th of July 2010 and shall be deemed to be effective as of July 23, 2010. \\\midrule
    \textsc{Effective Date Reference (91/111)} & This Contract shall become effective (the ``Effective Date'') upon the date this Contract is signed by both Parties. \\\midrule
    \textsc{No Solicitation (101/117)} Prohibition of acquiring employees of the other party (after the contract expires) and maintaining business relations with the customers of the other party. & You agree that for a~period of eighteen months (18) from the date hereof you will not directly or indirectly recruit, solicit or hire any regional or district managers, corporate office employee, member of senior management of the Company (including store managers), or other employee of the Company identified to you. \\\midrule
    \textsc{Confidential Information Form (152/174)} Forms and methods of providing confidential information. & ``Confidential Information'' means any technical or commercial information or data, trade secrets, know-how, etc., of either Party or their respective Affiliates whether or not marked or stamped as confidential, including without limitation, Technology, Invention(s), Intellectual Property Rights, Independent Technology and any samples of products, materials or formulations including, without limitation, the chemical identity and any properties or specifications related to the foregoing. Any Development Program Technology, MPM Work Product, MSC Work Product, Hybrid Work Product, Prior End-Use Work Product and/or Shared Development Program Technology shall be Confidential Information of the Party that owns the subject matter under the terms set forth in this Agreement. \\\midrule
    \textsc{Dispute Resolution (67/68)} Arrangements for how to resolve disputes (arbitration, courts). & The Parties will attempt in good faith to resolve any dispute or claim arising out of or in relation to this Agreement through negotiations between a~director of each of the Parties with authority to settle the relevant dispute. If the dispute cannot be settled amicably within fourteen (14) days from the date on which either Party has served written notice on the other of the dispute then the remaining provisions of this Clause shall apply. \\
    \bottomrule
    \caption{Clauses annotated in Non-disclosure Agreements. The values in parentheses indicate the number of documents with a~particular clause and the total number of clause instances, respectively.\label{tab:clauses1}}
    \end{longtable}
    \endgroup
}

\pagebreak

{\small
    \begingroup
    \begin{longtable}{p{0.37\textwidth} p{0.57\textwidth}}
    \toprule
    \textbf{Clause (Instances)} & \textbf{Example} \\
    \midrule
    \textsc{Change of Control Covenant (88/95)} Information about the obligation to redeem bonds for 101\% of the price in the event of change of control.  & Upon the occurrence of a~Change of Control Triggering Event (as defined below with respect to the notes of a~series), unless we have exercised our right to redeem the notes of such series as described above under ``Optional Redemption,''\noqa{latex-38} the indenture provides that each holder of notes of such series will have the right to require us to repurchase all or a~portion (equal to \$2,000 or an integral multiple of \$1,000 in excess thereof) of such holder's notes of such series pursuant to the offer described below (the ``Change of Control Offer''), at a~purchase price equal to 101\% of the principal amount thereof, plus accrued and unpaid interest, if any, to the date of repurchase, subject to the rights of holders of notes of such series on the relevant record date to receive interest due on the relevant interest payment date. \\\midrule
    \textsc{Change of Control Notice (78/79)} Information about the obligation to inform bondholders (usually by mail) about the event of change of control. This clause usually follows immediately the above clause. & Within 30 days following any Change of Control, B\&G Foods will mail a~notice to each holder describing the transaction or transactions that constitute the Change of Control and offering to repurchase notes on the Change of Control Payment Date specified in the notice, which date will be no earlier than 30 days and no later than 60 days from the date such notice is mailed, pursuant to the procedures required by the indenture and described in such notice. Holders electing to have a~note purchased pursuant to a~Change of Control Offer will be required to surrender the note, with the form entitled ``Option of Holder to Elect Purchase'' on the reverse of the note completed, to the paying agent at the address specified in the notice of Change of Control Offer prior to the close of business on the third business day prior to the Change of Control Payment Date. \\\midrule
    \textsc{Cross Default (96/110)} The company does not comply with certain conditions (event of default), so the bonds become due (e.g. when the company does not submit financial statements on time) — our clause was limited to the event of non-repayment, usually the minimum sum is given. & due to our default, we (i) are bound to repay prematurely indebtedness for borrowed moneys with a~total outstanding principal amount of \$75,000,000 (or its equivalent in any other currency or currencies) or greater, (ii) have defaulted in the repayment of any such indebtedness at the later of its maturity or the expiration of any applicable grace period or (iii) have failed to pay when properly called on to do so any guarantee of any such indebtedness, and in any such case the acceleration, default or failure to pay is not being contested in good faith and not cured within 15 days of such acceleration, default or failure to pay; \\\midrule
    \textsc{Litigation Default (42/51)} Court verdict or administrative decision which charge the company for a significant unpaid amount (another from the series of event of default). & (8) one or more judgments, orders or decrees of any court or regulatory or administrative agency of competent jurisdiction for the payment of money in excess of \$30 million (or its foreign currency equivalent) in each case, either individually or in the aggregate, shall be entered against the Company or any subsidiary of the Company or any of their respective properties and shall not be discharged and there shall have been a~period of 60 days after the date on which any period for appeal has expired and during which a~stay of enforcement of such judgment, order or decree, shall not be in effect; \\\midrule
    \textsc{Merger Restrictions (188/241)} A clause preventing the merger or sale of a company, etc., except under certain conditions (generally, the company should not avoid its obligations to its bondholders). & Without the consent of the holders of the outstanding debt securities under the indentures, we may consolidate with or merge into, or convey, transfer or lease our properties and assets to any person and may permit any person to consolidate with or merge into us. However, in such event, any successor person must be a~corporation, partnership, or trust organized and validly existing under the laws of any domestic jurisdiction and must assume our obligations on the debt securities and under the applicable indenture. We agree that after giving effect to the transaction, no event of default, and no event which, after notice or lapse of time or both, would become an event of default shall have occurred and be continuing and that certain other conditions are met; provided such provisions will not be applicable to the direct or indirect transfer of the stock, assets or liabilities of our subsidiaries to another of our direct or indirect subsidiaries. (Section 801) \\\midrule
    \textsc{Bondholders Default (191/241)} A clause on the payment of the principal amount and interest — they become due as a result of an event of default, if such a declaration is made by bondholders. & If an event of default (other than an event of default referred to in clause (5) above with respect to us) occurs and is continuing, the trustee or the holders of at least 25\% in aggregate principal amount of the outstanding notes by notice to us and the trustee may, and the trustee at the written request of such holders shall, declare the principal of and accrued and unpaid interest, if any, on all the notes to be due and payable. Upon such a~declaration, such principal and accrued and unpaid interest will be due and payable immediately. If an event of default referred to in clause (5) above occurs with respect to us and is continuing, the principal of and accrued and unpaid interest on all the notes will become and be immediately due and payable without any declaration or other act on the part of the trustee or any holders. \\\midrule
    \textsc{Tax Changes Call (48/56)} A clause about the possibility of an earlier redemption of the bond by the issuer if the tax law or its interpretation changes. & If, as a~result of any change in, or amendment to, the laws (or any regulations or rulings promulgated under the laws) of the Netherlands or the United States or any taxing authority thereof or therein, as applicable, or any change in, or amendments to, an official position regarding the application or interpretation of such laws, regulations or rulings, which change or amendment is announced or becomes effective on or after the date of the issuance of the notes, we become or, based upon a~written opinion of independent counsel selected by us, will become obligated to pay additional amounts as described above in ``Payment of additional \noqa{latex-38}amounts,'' then the Issuer may redeem the notes, in whole, but not in part, at 100\% of the principal amount thereof together with unpaid interest as described in the accompanying prospectus under the caption ``Description of WPC Finance Debt Securities and the Guarantee-Redemption for Tax Reasons.'' \\\midrule
    \textsc{Financial Statements (201/317)} A clause on the obligation to submit (usually to the SEC) annual reports or other reports. & Notwithstanding that the Company may not be subject to the reporting requirements of Section 13 or 15(d) of the Exchange Act, the Company will file with the SEC and provide the Trustee and Holders and prospective Holders (upon request) within 15 days after it files them with the SEC, copies of its annual report and the information, documents and other reports that are specified in Sections 13 and 15(d) of the Exchange Act. In addition, the Company shall furnish to the Trustee and the Holders, promptly upon their becoming available, copies of the annual report to shareholders and any other information provided by the Company to its public shareholders generally. The Company also will comply with the other provisions of Section 314(a) of the TIA. \\
    \bottomrule
    \caption{Clauses annotated in Corporate Bonds. The values in parentheses indicate the number of documents with a~particular clause and the total number of clause instances, respectively.\label{tab:clauses2}}
    \end{longtable}
    \endgroup
}

\twocolumn


\end{document}